\documentclass{bvm2022} 

\begin{document}



\selectlanguage{english} 

\title{Deep learning-based Subtyping of Atypical and Normal Mitoses using a Hierarchical Anchor-Free Object Detector}

\titlerunning{Subtyping of Atypical and Normal Mitosis}

\author{
	Marc \lname{Aubreville} \inst{1},
    Jonathan \lname{Ganz} \inst{1},
    Jonas \lname{Ammeling} \inst{1},
	Taryn A. \lname{Donovan} \inst{2}, \\
    Rutger H. J. \lname{Fick} \inst{3},
	Katharina \lname{Breininger} \inst{4},
	Christof A. \lname{Bertram} \inst{5}
}

\authorrunning{Aubreville et al.}

\institute{
\inst{1} Technische Hochschule Ingolstadt, Ingolstadt, Germany\\
\inst{2} Schwarzman Animal Medical Center, New York, NY, USA\\
\inst{3} Tribun Health, Paris, France\\ 
\inst{4} Department AIBE, Friedrich-Alexander-Universit\"at Erlangen-N\"urnberg, Germany\\
\inst{5} Institute of Pathology, University of Veterinary Medicine Vienna, Vienna, Austria\\
}

\email{marc.aubreville@thi.de}

\maketitle

\begin{abstract}
Mitotic activity is key for the assessment of malignancy in many tumors. Moreover, it has been demonstrated that the proportion of abnormal mitosis to normal mitosis is of prognostic significance. Atypical mitotic figures (MF) can be identified morphologically as having segregation abnormalities of the chromatids.
In this work, we perform, for the first time, automatic subtyping of mitotic figures into normal and atypical categories according to characteristic morphological appearances of the different phases of mitosis. Using the publicly available MIDOG21 and TUPAC16 breast cancer mitosis datasets, two experts blindly subtyped mitotic figures into five morphological categories. Further, we set up a state-of-the-art object detection pipeline extending the anchor-free FCOS approach with a gated hierarchical subclassification branch. 
Our labeling experiment indicated that subtyping of mitotic figures is a challenging task and prone to inter-rater disagreement, which we found in 24.89\% of MF. Using the more diverse MIDOG21 dataset for training and TUPAC16 for testing, we reached a mean overall average precision score of 0.552, a ROC AUC score of 0.833 for atypical/normal MF and a mean class-averaged ROC-AUC score of 0.977 for discriminating the different phases of cells undergoing mitosis. 

\end{abstract}

\section{Introduction}
The significance of identifying and counting mitotic figures in histopathology samples has been demonstrated for many tumor types \cite{donovan2021mitotic}. Errors of chromosome segregation that occur during cell division can appear morphologically as atypical mitotic figures (AMFs) and may correlate with genetic abnormalities \cite{lashen2022characteristics}. AMFs can be placed into two categories including mitotic/polar asymmetry and abnormal segregation of sister chromatids \cite{donovan2021mitotic}. Moreover, it has been long suspected \cite{rubio1991atypical} and recently confirmed \cite{ohashi2018prognostic,lashen2022characteristics,jin2007distinct} that not only the number of mitotic figures per unit area (the mitotic count) is of prognostic significance, but that atypical morphology is also prognostically important. Ohashi \etal\, were the first to investigate the significance of AMFs in breast cancer, and found that the presence of chromosome lagging and spindle multipolarity had a higher prognostic value than the mitotic figure count or Ki-67 immunohistochemistry-stained cell count \cite{ohashi2018prognostic}. In a larger study by Lashen \etal\, the authors found that a high atypical-to-typical mitotic figure ratio is associated with poor breast cancer-specific survival and is a predictor of poor response to chemotherapy in triple negative breast cancer \cite{lashen2022characteristics}.  

The detection and counting of mitotic figures on digital histopathology images stained with standard hematoxylin and eosin is a task that can be automated using deep learning architectures \cite{aubreville2022mitosis}. These advances are possible because of the availability of large-scale, diverse datasets such as the TUPAC16 \cite{veta2019predicting} and MIDOG21 challenge \cite{aubreville2022mitosis} datasets along with the availability of modern deep learning-based object detection pipelines. Yet, there are currently no automated pipelines reported in literature that have considered detection of mitotic figures and subclassification according to their morphology into normal and atypical mitotic figures or into the various phases within the mitosis. Automated approaches are therefore necessary for investigating these aspects on a larger scale. 
Distinct types of abnormalities may occur in the different mitosis phases and separation of the normal mitotic figure morphologies along the cell cycle may facilitate identification of AMF. Further, 
identification of mitosis phase might be a valuable tool for high throughput cell analysis within the development of novel cancer targets \cite{tao2007support}. 

We propose an efficient and fully automatic pipeline based upon the principle of Fully Convolutional One-Stage Object Detection (FCOS) \cite{tian2019fcos}. Considering the different phases of mitotic figures and the morphological variations manifested as AMF, the classes of a one-dimensional classification problem does not account for the hierarchy of classes. Thus, we extend the architecture by a hierarchical class prediction and define a tailored loss function and sampling scheme. 

\begin{figure}[t]
\centering
\includegraphics[width=0.7\linewidth]{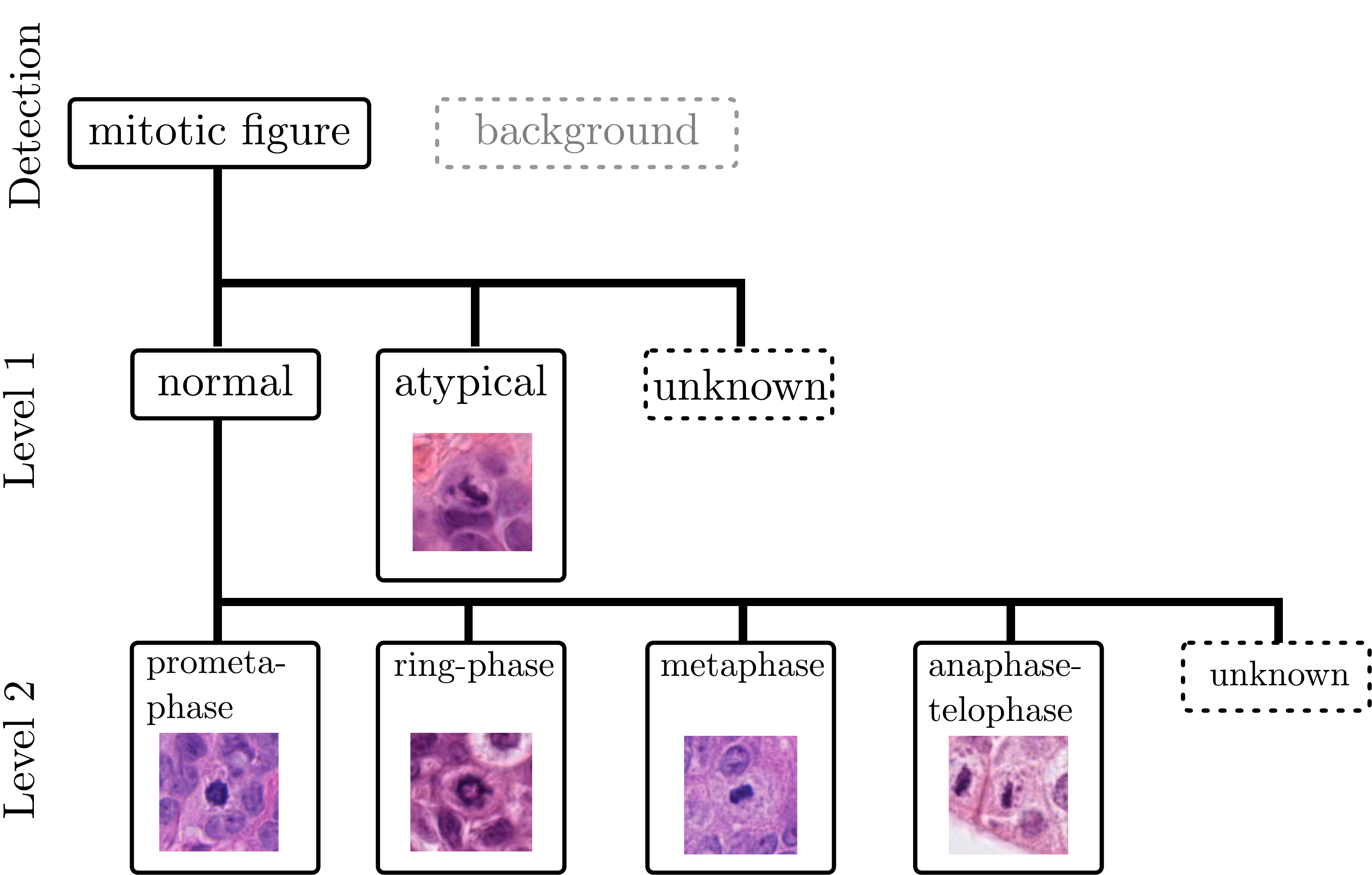}
\caption{Label hierarchy: We consider the detection and the subclassification of mitotic figures in two levels.}
\label{fig:subclasses}
\end{figure}

\section{Dataset}
All mitotic figures contained within the TUPAC16 \cite{veta2019predicting} and MIDOG21 \cite{aubreville2022mitosis} datasets were annotated by two pathologists according to five classes, spread across two hierarchical levels (Fig. \ref{fig:subclasses} and Tab. \ref{tab:annotations}): On level 1, we differentiated between atypical and normal mitotic figures. The latter was subsequently split on level 2 into prometaphase, metaphase, ring shaped mitosis, and anaphase/telophase. The category of atypical mitotic figures was not subdivided further due to the limited prevalence. For the annotation, we used the definitions used by Donovan \etal\ \cite{donovan2021mitotic}.
We used the alternative version of the TUPAC16 dataset \cite{bertram2020pathologist} provided by the same authors as the MIDOG21 dataset to reduce potential label bias. For annotations, we exported all mitotic figures as image cropouts with sufficient contextual information and asked two pathologists independently to assign the respective subcategories for both hierarchy levels. We found a total agreement in 75.11\% of all cells.  As shown in Table \ref{tab:annotations}, the agreement was the highest for normal mitotic figures, while the the lack of agreement was most significant between atypical mitotic figures and prometaphase and metaphase mitotic figures. 

\renewcommand{\arraystretch}{1.25}
\setlength{\tabcolsep}{5pt}

\begin{table}[t]
    \centering
    \resizebox{\textwidth}{!}{
\begin{tabular}{ll|r|rrrr}
		\hline
 &  & atypical  &  \multicolumn{4}{c}{typical} \\
  &  & &  ring shape  &  anaphase-telophase &  prometaphase &  metaphase \\

\hline
atypical &                                    &  587 &                                       6 &                         17 &                   39 &                79 \\
\hline
\multirow{4}{*}{typical} & ring shape (prometaphase or metaphase) &   21 &                                      57 &                          2 &                    3 &                 3 \\
& normal anaphase-telophase              &   88 &                                       0 &                        218 &                    3 &                17 \\
& normal prometaphase                    &  258 &                                     128 &                         25 &                 1200 &               280 \\
& normal metaphase                       &  176 &                                       0 &                         12 &                   30 &              1520 \\

\hline
\end{tabular} 
}
\caption{Confusion matrix of expert 1 (columns) and expert 2 (rows) in our annotation experiment on the TUPAC16 and MIDOG21 datasets (total number of objects was 4,769).}
    \label{tab:annotations}
\end{table}

\section{Methods}
The expert disagreement in almost a quarter of cases raises the question of how to deal with the expert disagreement. Particularly for the four-category discrimination on our secondary level (mitotic phase), incorporating the opinion of a blinded tertiary expert would not necessarily lead to a clear majority vote.  Instead, our approach aimed at mitigating this disagreement: We formulate the object detection problem as a hierarchical label problem with the main object class (mitosis) being a categorical label that a majority of experts agreed upon. We introduce two subcategorical labels for each object that can be either present (in case of an unanimous vote of the experts) or absent (Fig. \ref{tab:annotations}). 

\subsection{Deep Learning Architecture}
\begin{figure}[t]
    \centering
    \includegraphics[width=\linewidth]{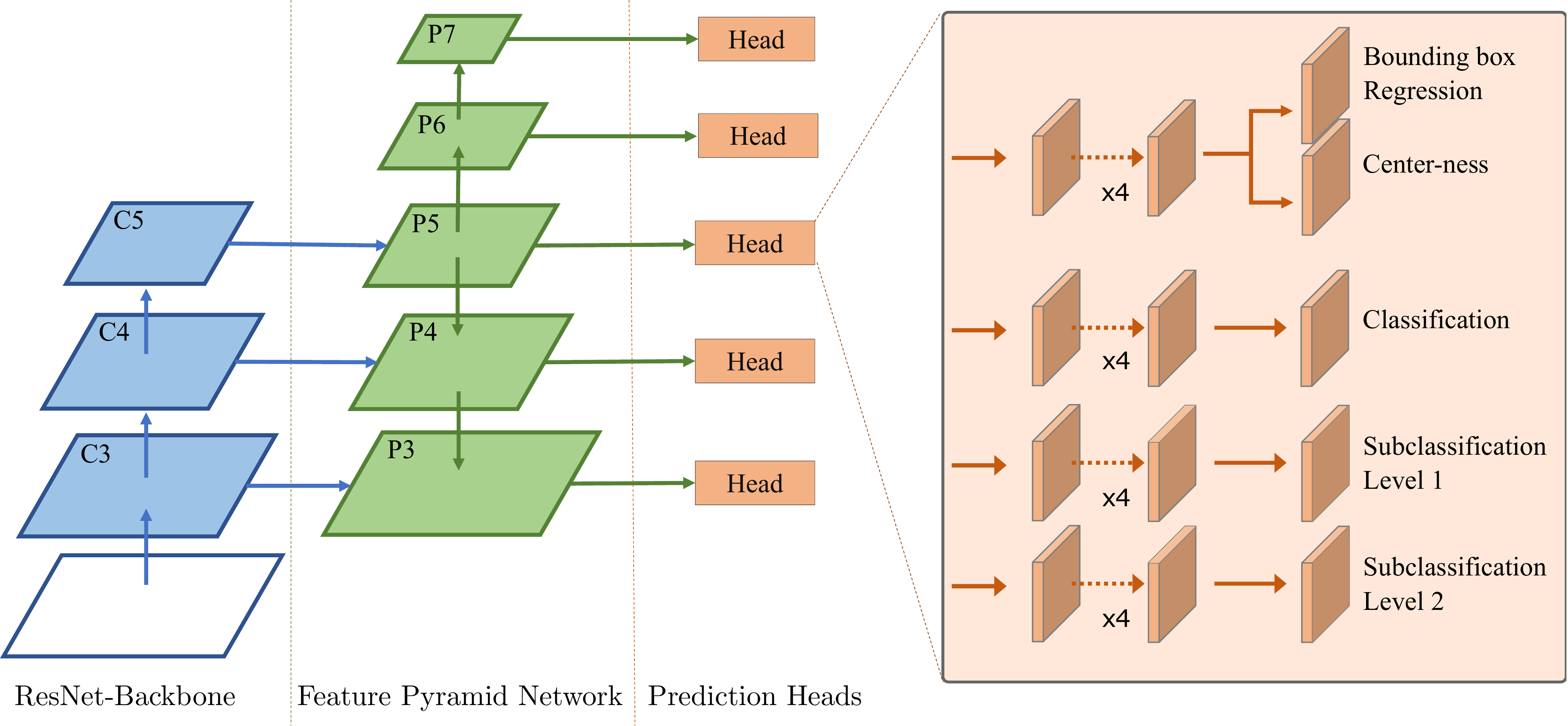}
    \caption{Depiction of the model architecture, with two additional hierarchical label classification heads. Modified from \cite{tian2019fcos}. }
    \label{fig:architecture}
\end{figure}

We based our experiments on the FCOS architecture by Tian \etal\cite{tian2019fcos}. Over previous approaches, FCOS does not depend on the prior definition and optimization of anchor boxes for object detection and instead predicts object classes and bounding boxes as well as the distance to the object center for each coordinate in the latent space of a feature pyramid network. As shown in Figure \ref{fig:architecture}, we extended the architecture by two new heads, corresponding to the two hierarchical levels of our problem. 
Accounting for the significant disagreement between experts that Tab. \ref{tab:annotations} reveals, we added also an \textit{unknown} class on each hierarchy level to account for expert disagreement. For both new subclassification heads, we added a loss-term to the FCOS loss, realized as a generalized intersection over union-dependent focal loss as in FCOS, but gated with the availability of the subclass, i.e., zeroed for object-related losses where the respective subclass is \textit{unknown}. As a baseline, we used an unmodified, non-hierarchical FCOS approach with 6 classes: mitotic figure with unknown subclass, atypical, and the four normal classes. In both cases, we used square patches of size 1024. All prediction heads had 4 CNN layers, as in the default FCOS. We make all code available online\footnote{\url{https://github.com/DeepPathology/HierarchicalFCOS/}}. 

\subsection{Training}

We trained and validated both models on the MIDOG21 dataset and used the less diverse (particularly regarding the scanner variability) TUPAC16 dataset as our hold-out test set. To account for variability in training, especially related to a potential domain shift between datasets, we repeated the experiments five times. 
We randomly assigned 20\% of the MIDOG21 cases to the validation set, which we used for model selection and used the remainder of the training set for the optimization of the model weights. 
We stratified for the six classes within our sampling scheme to make sure a sufficient gradient accumulation for the classes of low occurrence. Our sampling scheme ensured that at least one mitotic figure of the selected class would be visible on the selected image patch.
Additionally, we employed sampling of hard negatives examples in 30\% of cases, which aims to make the model more robust (i.e., able to differentiate mitotic figures from mitotic figure look-alikes). The training dataset provides hard examples that can be used for this purpose.
We defined an epoch as containing 2000 randomly sampled images. We then trained for up to 50 epochs, using early stopping on the validation loss with a patience of 10. 

\subsection{Evaluation}

Average precision (AP) is commonly calculated class-specific for each of the object detection classes, based upon the score that was assigned to that object, and averaged to yield the mean average precision (mAP) score. However, having only a single class (mitosis) in our hierarchical label approach and multiple classes in the baseline, this makes the mAP metric for this case practically incomparable. Instead, we calculated for both models the overall (class-independent) AP score. For the baseline model this was achieved by ignoring the class of the object for both, the labels and the detections. Similarly, we report the overall average recall (AR).
Additionally, we report the area under the receiver operating characteristic curve (ROC-AUC) of all objects that were detected by the respective approaches (using the default threshold of 0.5 for the intersection over union). For the multi-class problem in hierarchy level two, we report the ROC-AUC as one-vs-rest classification problem. 

\section{Results}

\begin{table}[]
    \centering
    \resizebox{\textwidth}{!}{
    \begin{tabular}{l|c|cccc|c|c}
      approach & ROC-AUC Level 1  & \multicolumn{4}{c|}{ROC-AUC Level 2} & Overall AP & Overall AR \\
      &  atypical/normal & prometaphase & ring shape & metaphase & anaphase-telophase & \\
\hline
FCOS (baseline)& $0.817 \pm 0.019$& $0.957 \pm 0.024$& $0.926 \pm 0.011$& $0.924 \pm 0.009$& $0.945 \pm 0.008$& $0.499 \pm 0.061$& $0.929 \pm 0.012$\\
HFCOS (proposed)& $\mathbf{0.833} \pm 0.061$&$\mathbf{0.971} \pm 0.023$&$\mathbf{0.966} \pm 0.023$&$\mathbf{0.984} \pm 0.012$&$\mathbf{0.987} \pm 0.003$& $\mathbf{0.552} \pm 0.084$& $\mathbf{0.946} \pm 0.013$\\
\end{tabular}}
    \caption{Results of the proposed hierarchical anchor-free model in comparison to a baseline non-hierarchical baseline model. Given are mean $\pm$ std values of the area under the receiver operating characteristic curve (ROC-AUC) of detected objects of five consecutive runs as well as the overall average precision (AP) and average recall (AR).}
    \label{tab:results}
\end{table}

We found that the subclassification benefits from the hierarchical approach (Tab. \ref{tab:results}). We observed this effect both for the level 1 subclassification of atypical vs. normal mitotic figures and for the level 2 subclassification of mitotic phases. Effectively, for all observed metrics, the proposed approach outperformed the baseline. Moreover, we found a moderate variability of the results across the five runs. Further, assessing the overall cell detection, we found the AP and AR for mitotic figures overall to be increased by using the hierarchical labels model. 

\section{Discussion}

Our results, particularly those of the annotation experiment (Tab. \ref{tab:annotations}), indicate that differentiation between normal and atypical mitotic figures (AMFs) represents a difficult task. However, identification of the phases of mitosis (when clearly defined), is a more straightforward task for humans and deep learning algorithms. It is possible that a refined definition of the features of AMFs may improve agreement between experts, or that some of the subcategories of AMFs may have better agreement than others. The overall moderate AP value could be linked to a domain shift between the MIDOG21 and the TUPAC16 data set, and could be reduced by applying methods of domain generalization~\cite{aubreville2022mitosis}, and/or tackled by more extensive data augmentation. 
One limitation of the current work is that the MIDOG21 and the TUPAC16 datasets include only regions from the areas of increased mitotic activity (hotspots) within the breast cancer specimens. Thus, it is unclear how robust the trained algorithm would perform on whole slide images (WSIs), and clinical-grade solutions would need to be trained on combined data sets including fully annotated WSI. Furthermore, the MIDOG21 dataset uses data from only one hospital, and generalization might also be reduced by this.
To the best of our knowledge, this approach represents the first attempt at automated mitotic figure detection and subclassification on H\&E-stained tissue, both for the identification of AMFs and the mitotic phase for normal mitoses. The availability of this method enables extensive clinical research investigating the prognostic power of the ratio of AMFs to normal mitotic figures, which we will continue to expand upon in our ongoing research.

\printbibliography

\end{document}